\documentclass[conference,a4paper]{IEEEtran}
\IEEEoverridecommandlockouts

\usepackage[hidelinks]{hyperref}
\usepackage[cmex10]{amsmath}
\usepackage{amssymb,amsfonts}
\usepackage{dblfloatfix}

\usepackage[ruled,vlined]{algorithm2e}
\usepackage{graphicx}
\graphicspath{{Figures/PDF/}{Figures/PNG/}}

\usepackage{booktabs}
\usepackage{siunitx}
\usepackage[numbers,compress]{natbib}
\usepackage{texnames}
\usepackage{bm,bbm}
\usepackage{orcidlink}
\usepackage{float}
\usepackage{graphicx}
\usepackage{longtable}
\usepackage{subfigure}
\usepackage{dblfloatfix}

\begin{document}

\title{\uppercase{Monitoring digestate application on agricultural crops using Sentinel-2 Satellite imagery}
\thanks{\textcopyright\ 2025 IEEE. Published in the 2025 IEEE International Geoscience and Remote Sensing Symposium (IGARSS 2025), scheduled for 3 - 8 August 2025 in Brisbane, Australia. Personal use of this material is permitted. However, permission to reprint/republish this material for advertising or promotional purposes or for creating new collective works for resale or redistribution to servers or lists, or to reuse any copyrighted component of this work in other works, must be obtained from the IEEE. Contact: Manager, Copyrights and Permissions / IEEE Service Center / 445 Hoes Lane / P.O. Box 1331 / Piscataway, NJ 08855-1331, USA. Telephone: + Intl. 908-562-3966.}
\thanks{This version is the accepted manuscript submitted to arXiv. The final version will be published in the Proceedings of IGARSS 2025 and available via IEEE Xplore. For citation, please refer to the published version in IGARSS 2025.}
}

\author{
    Andreas Kalogeras\textsuperscript{1},
    Dimitrios Bormpoudakis\textsuperscript{1},
    Iason Tsardanidis*\textsuperscript{1},     \thanks{*Correspondence to  \href{mailto:j.tsardanidis@noa.gr}{j.tsardanidis@noa.gr}}
    Dimitra A. Loka\textsuperscript{2}
    and Charalampos Kontoes\textsuperscript{1}\\[1em]
    
    \textsuperscript{1}BEYOND EO Centre, IAASARS, National Observatory of Athens, Athens, Greece\\
    \textsuperscript{2}Institute of Industrial and Forage Crops, Hellenic Agricultural Organization (ELGO) “DIMITRA”, Larissa, Greece
    
}

    


\maketitle
\begin{abstract}
The widespread use of Exogenous Organic Matter in agriculture necessitates monitoring to assess its effects on soil and crop health. This study evaluates optical Sentinel-2 satellite imagery for detecting digestate application, a practice that enhances soil fertility but poses environmental risks like microplastic contamination and nitrogen losses. In the first instance, Sentinel-2 satellite image  time series (SITS) analysis of specific indices (EOMI, NDVI, EVI) was used to characterize EOM's spectral behavior after application on the soils of four different crop types in Thessaly, Greece. Furthermore,  Machine Learning (ML) models (namely Random Forest, k-NN, Gradient Boosting and a Feed-Forward Neural Network), were used to investigate digestate presence detection, achieving F1-scores up to 0.85. The findings highlight the potential of combining remote sensing and ML for scalable and cost-effective monitoring of EOM applications, supporting precision agriculture and sustainability.

\end{abstract}


\section{Introduction}



Agricultural systems can benefit from the application of Exogenous Organic Matter (EOM), which not only enhances soil fertility but also supports waste recycling and promotes circular economies \cite{Dodin31122023}, \cite{CHEN2022105415}. One notable source of EOM is digestate, a byproduct of anaerobic digestion—a process which plays a significant role in biogas production \cite{SCARLAT2018457}. The characteristics of digestate vary depending on the feedstock and production conditions \cite{GUILAYN201967}, and its application to soils.


Mapping digestate application in real-world conditions can be leveraged to monitor policy compliance (e.g., in the case of subsidies) or to enable large-scale assessments of its effects on crucial agro-ecological parameters such as crop productivity, soil organic matter, or soil microbial communities \cite{TANG2006200}. Nevertheless, detecting digestate application presents significant challenges. While farmer or government monitoring agency surveys are common, remote sensing is a promising alternative, having been used to monitor agricultural practices
\cite{tsardanidis2024}
, irrigation needs \cite{electronics12010127}, and crop conditions \cite{sitokonstantinou2023fuzzy}. 

However, the mapping of EOM application through remote sensing remains underexplored. Laboratory studies have examined EOM types like grape marc compost and cattle manure compost \cite{BENDOR19971}, hog manure \cite{article}, and poultry manures \cite{doi:10.1177/09670335211007543}. Sentinel-2 satellite image time series (SITS) have been used to monitor EOM applications \cite{rs13091616}, \cite{SHEA2022115334}, \cite{PEDRAYES2023102006}, linking spectral measurements to imagery, but in relation to digestate at limited and controlled test field scales \cite{rs13091616}. Here, we evaluate the potential of remote sensing for detecting digestate application as it is applied in a large-scale, practical, real-world situation.

EOM shares low reflectance characteristics with soil organic carbon in visible wavelengths \cite{JOURAIPHY2005101}, \cite{rs13091616}, enabling spectral indices calculated using Sentinel-2 red (B04), near-infrared (NIR, B08), and shortwave infrared (SWIR, B11 and B12) bands to detect its application under certain conditions. Composed of non-living tissues, EOM differs spectrally from vegetation, which strongly absorbs red and blue wavelengths and reflects in infrared \cite{verhoeven2006looking}. Liquid EOM applied to active vegetation has been shown to be detectable shortly after application, though crop growth and drying effects remain unexplored, highlighting the need for more research.

This study aims to assess the effectiveness of Sentinel-2 imagery in monitoring EOM application in the field, with a focus on digestate implementation in agricultural parcels of different crop types. By integrating multispectral satellite data to capture EOM-induced changes in soils and vegetation, the study seeks to develop scalable, transferable methodologies for effective monitoring across diverse crop types.

\section{Materials and Methods}
\subsection{Study site and field description}

\begin{figure}[H]
    \centering
    \includegraphics[width=0.95\linewidth]{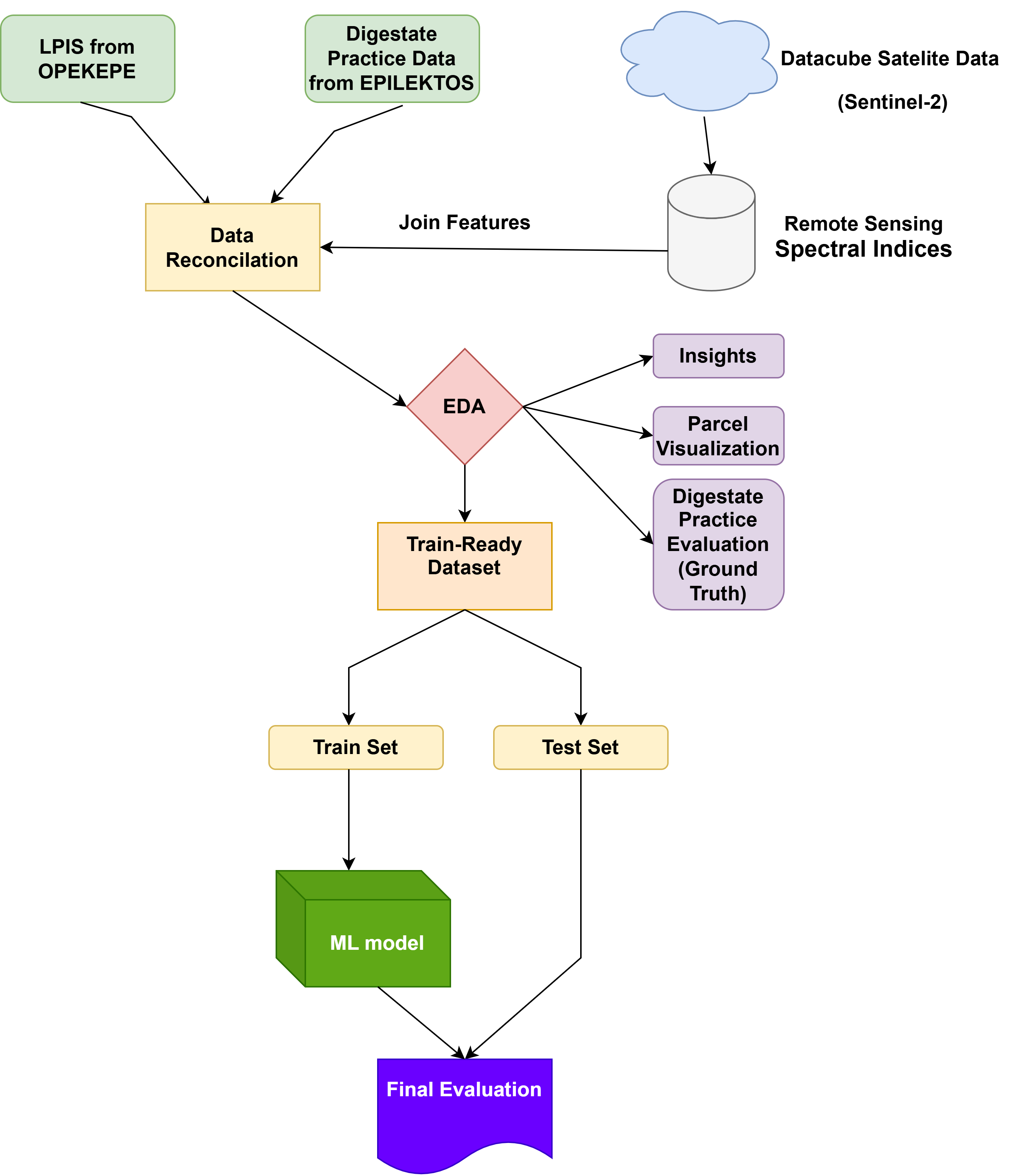}
    \caption{Pipeline of the process}
    \label{fig: Fig.1}
\end{figure}

For the development of the experiments, Land Parcel Identification System (LPIS) data for 2023, including crop type and parcel geometry, were provided by the Greek Payment Authority "OPEKEPE" for the EU's Common Agricultural Policy (CAP) subsidies. These agricultural parcels are located in the region of Thessaly, Greece. Information on digestate application events, location, date and quantity of application was supplied by "EPILEKTOS" SELECTED TEXTILE INDUSTRIAL COMPANY S.A. 
The initial dataset comprised 272 parcels, of which 97 underwent treatment, while the remaining parcels served as controls.
The general approach, as illustrated in Fig. \ref{fig: Fig.1}, involves analyzing the temporal changes in the spectral behavior of partially vegetated soils treated with EOM. This analysis was conducted using Sentinel-2 imagery. To ensure the temporal alignment of digestate application and satellite data, Sentinel-2 images were selected based on their proximity to the dates of application. Sentinel-2 data were extracted for each parcel. The Scene Classification Map (SCL) generated by Sen2Cor \cite{sen2cor_new} atmospheric correction was filtered to include only values corresponding to vegetation, non-vegetated areas, and water. Additionally, a cloud coverage threshold of 20\% was applied.

\subsection{Spectral indices for EOM detection}

In total, seven spectral indices were computed. $EOMI_1$, $EOMI_2$, $EOMI_3$, $EOMI_4$ and $NBR_2$ (Eq. 1-5) were previously proposed to monitor digestate application. $EOMI_3$ (Eq.3) combines $EOMI_1$ and $EOMI_2$ and uses the red, NIR and SWIR bands of Sentinel-2 most impacted by solid EOM. In order to consider winter wheat vegetation, $EVI$ and $NDVI$ have also been used. Finally, the ratios between EOM and vegetation indices were calculated to account for the combined effects of growing vegetation and EOM application.

\begin{equation}\label{eq:EOMI1}
EOMI_1 = \frac{B11 - B8A}{B11 + B8A}
\end{equation}
\begin{equation}\label{eq:EOMI2}
EOMI_2 = \frac{B12 - B04}{B12 + B04}
\end{equation}
\begin{equation}\label{eq:EOMI3}
EOMI_3 = \frac{(B11 - B8A) + (B12 - B04)}{(B11 + B8A) + (B12 + B04)}
\end{equation}
\begin{equation}\label{eq:EOMI4}
EOMI_4 = \frac{B11 - B04}{B11 + B04}
\end{equation}
\begin{equation}\label{eq:NBR_2}
NBR_2 = \frac{B11 - B12}{B11 + B12}
\end{equation}
\begin{equation}\label{eq:NDVI}
NDVI = \frac{B08 - B04}{B08 + B04}
\end{equation}
\begin{equation}\label{eq:EVI}
EVI = 2.5 \cdot \frac{B08 - B04}{B08 + 6 \cdot B04 - 7.5 \cdot B02 + 1}
\end{equation}

where $B02$, $B04$, $B08$, $B8A$, $B11$ and $B12$ refer to the Sentinel-2 spectral bands.

Spectral indices were computed across all parcels in the dataset for all available Sentinel-2 imagery, extended 30 days before and 30 days after the application of digestate,
since in several cases, the implementation was carried out in multiple stages over a short time frame.


\begin{figure*}[ht!]
    \centering
    \subfigure[ ]
    {
        \includegraphics[width=0.8\linewidth]{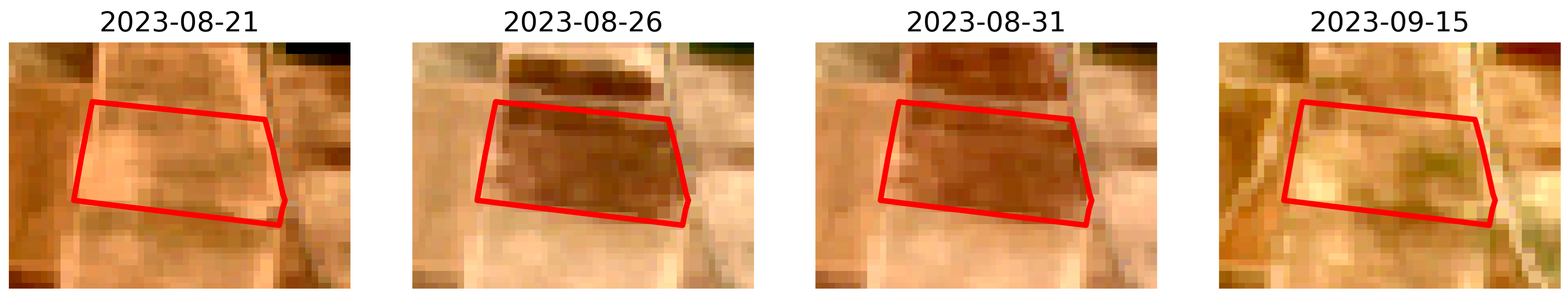}
        \label{fig:first_sub_1}
    }
    \\
    \subfigure[ ]
    {
        \includegraphics[width=0.8\linewidth]{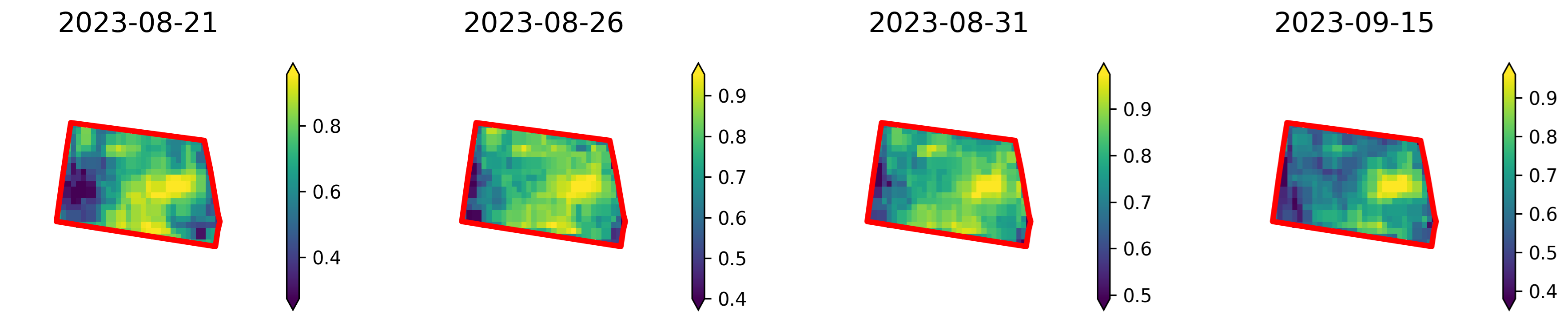}
        \label{fig:second_sub_1}
    }
    \subfigure[ ]
    {
        \includegraphics[width=0.8\linewidth]{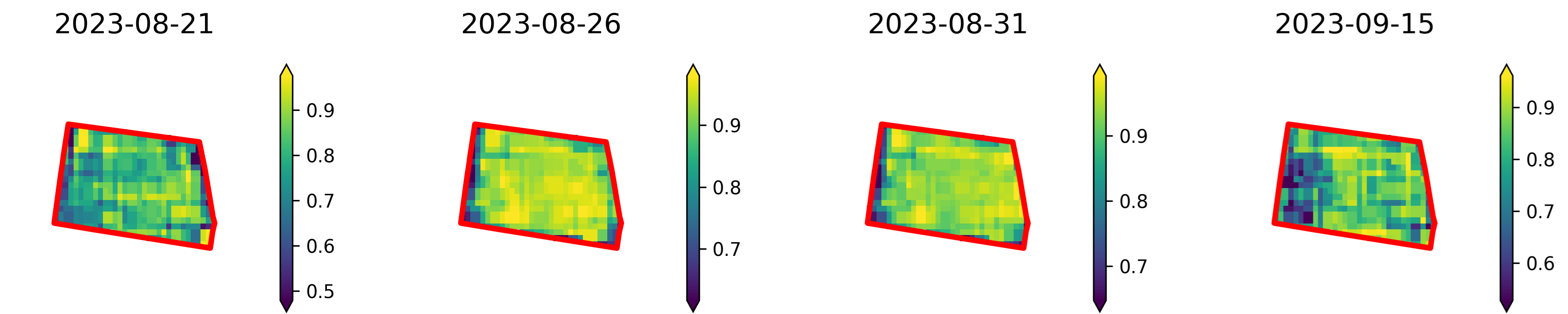}
        \label{fig:third_sub_1}
    }
    \caption{Comparison of three monitoring approaches for assessing field conditions after digestate application: (a) standard RGB imagery, (b) vegetation health using the NDVI, and (c) exogenous organic matter using the EOMI2.}
    \label{fig:combined-vertical}
\end{figure*}

\subsection{Model Implementation}
The main goal of this study is to use ML to identify whether an agricultural parcel has received digestate treatment. "EPILEKTOS S.A."' data annotates whether a parcel has undergone treatment or not, as it contains information regarding the treatment date(s) and the quantity of EOM per field. We labeled the data with 0 for untreated fields and 1 for treated fields. The ML models were then trained to predict the correct label based on these events (Binary Classification problem).

Following feature engineering and the application of one-hot encoding to categorical labels (crop type), the dataset was randomly split into training and test sets using stratified sampling in an 80/20 ratio. Three ML models were selected for training: Random Forest, Gradient Boosting, and k-Nearest Neighbors. The hyperparameters for the models were selected as follows: for the Random Forest, 100 estimators were used with no maximum depth and the square root of the number of features as the maximum number of features. For k-Nearest Neighbors, 5 neighbors were chosen with uniform weights. For Gradient Boosting, 100 estimators were utilized with a learning rate of 0.1. Furthermore, a feed-forward neural network was constructed that consists of an input layer that receives the preprocessed features, followed by a single hidden layer with 128 neurons and a ReLU activation function to introduce non-linearity. The output layer uses a Softmax activation function to produce class probabilities. This architecture is trained using Cross-Entropy Loss and optimized with the Adam optimizer, which adapts the learning rate during training. The network is trained over 10 epochs. For all models, evaluation was conducted using a 5-fold cross-validation strategy based on Precision, Recall, and F1 Score statistics.


\section{Results and Discussion}

\subsection{Photo Interpretation}

\begin{figure}[ht!]
    \centering
    \subfigure[ ]
    {
        \includegraphics[width=\linewidth]{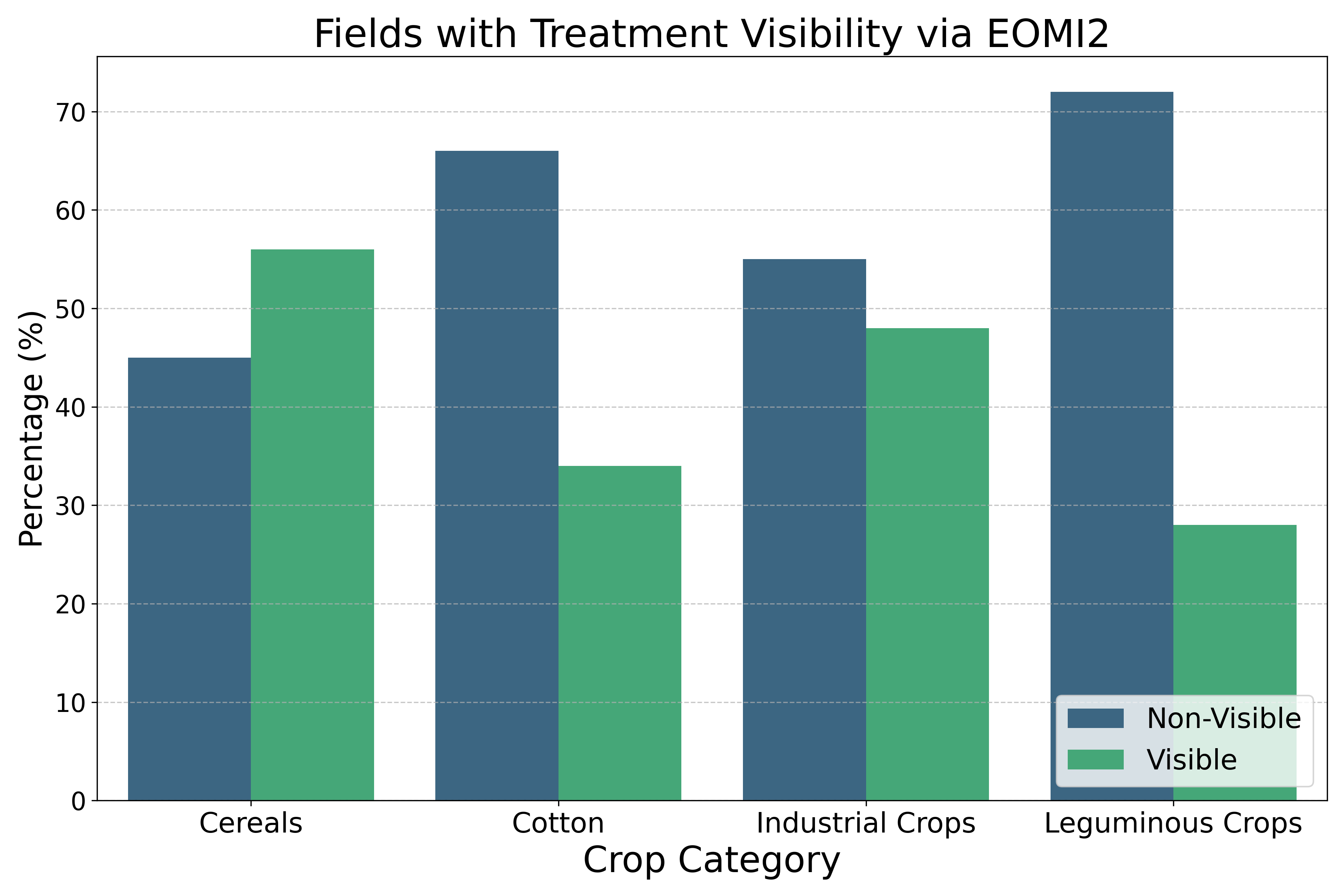}
        \label{fig:first_sub_2}
    }
    \\
    \subfigure[ ]
    {
        \includegraphics[width=\linewidth]{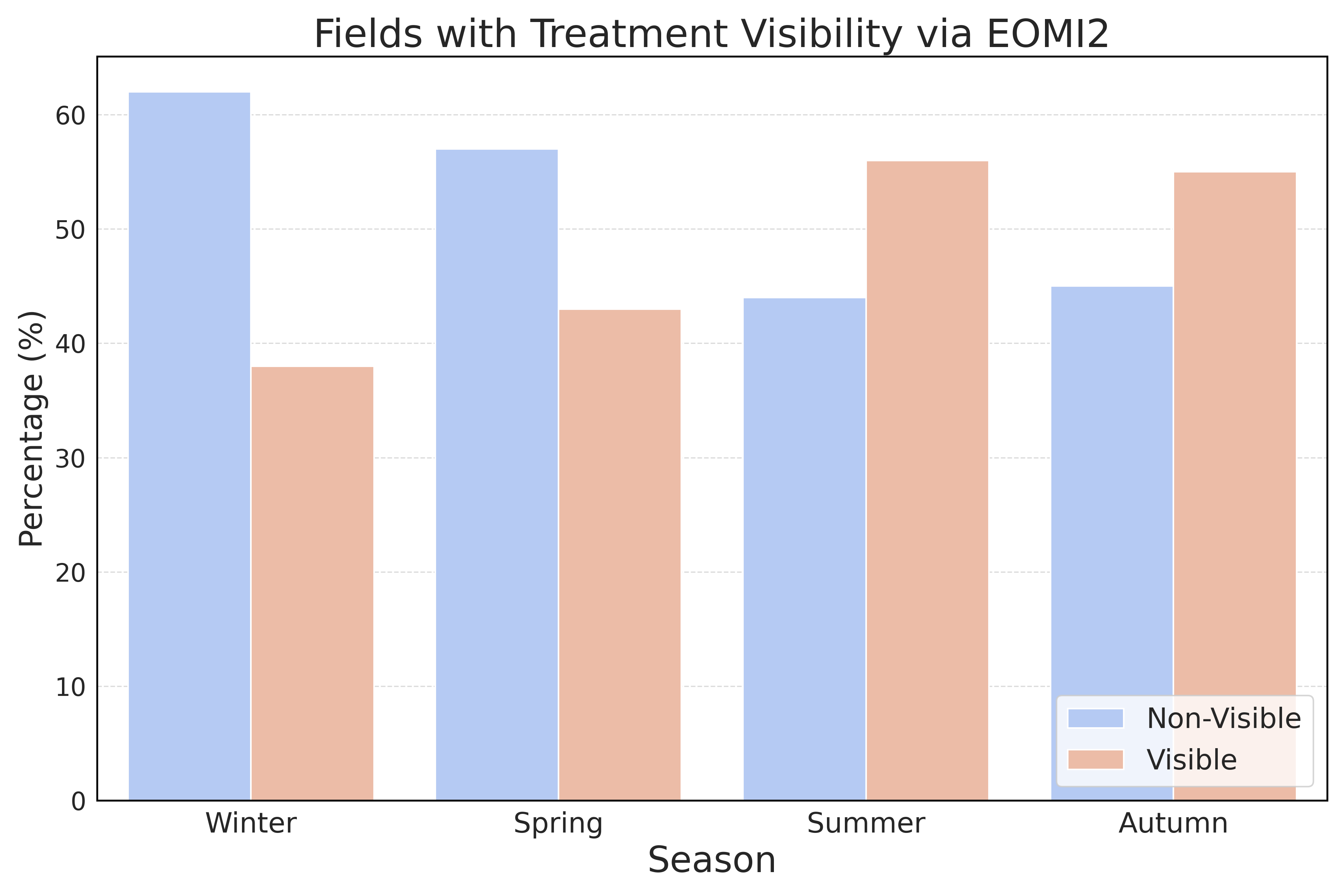}
        \label{fig:second_sub_2}
    }
    \caption{Comparison of barcharts showing the percentages of appearance or not of the change in the index: (a) per crop category, and (b) per season.}
    \label{fig:combined-horizontal}
\end{figure}

After constructing the dataset, the results of photo interpretation within the RGB scope, $NDVI$ and $EOMI_2$ reveal distinct insights into the field conditions.
We first visualized a parcel where the application of digestate is detectable through changes in the $EOMI_2$ index, and compared this to its RGB image and $NDVI$ values (see Fig. \ref{fig:combined-vertical}). This particular parcel had several dates of digestate application, with the most recent being on 28-08-2023, and a significant amount of EOM was applied. Notably, we observed visible changes even in the RGB image. However, an interesting finding was that the $NDVI$ index showed minimal, if any, changes following the digestate application, in contrast to the $EOMI_2$, which clearly detected the changes. 
After conducting the photo interpretation for all parcels, we found that 48 out of 97 fields did not exhibit a specific change in the $EOMI_2$ index. Therefore, the recall of detecting the change solely through photo interpretation is 50.51\%.
Additionally, we collected the crop codes for the agricultural parcels to determine whether there is any relationship between the crop type and the appearance of visible change in the photo interpretation. Since many crop codes belong to the same broader family, we grouped the crops into four main categories: Cereals, Cotton, Industrial Crops, and Leguminous Crops. Fig. \ref{fig:combined-horizontal} shows how annotations of actual changes in the $EOMI_2$ index are distributed across these categories and seasons.
The plot reveals that the application of digestate is more visible for some crop types than others. However, in most cases, the distribution is nearly equal. As a result, no definitive conclusion can be drawn from visual inspection of spectral indices regarding the relationship between crop type and digestate application. The same can be observed in the second plot, where the comparison is implemented per season.





\subsection{Model Evaluation}
We implemented various ML models to assess the effectiveness of Sentinel-2 imagery in monitoring EOM application, with a focus on digestate implementation in agricultural fields of different crop types, and the results are diplayed in Table 1.
The Random Forest model excels in identifying the negative class (class 0), showing high recall and a decent F1-score for this class. However, its ability to correctly identify the positive class (class 1) is limited, with a significant drop in recall. This suggests that the model may be overly biased toward the majority class (class 0).
A similar observation can be made for k-Nearest Neighbors and Gradient Boosting, although they score slightly lower(significantly lower in class 1 precision), with k-Nearest Neighbors showing a more notable decrease in performance.
The performance of the Feed-Forward Neural Network is between Gradient Boosting and Random Forest, having the best score in recall.

\begin{table}[H]
\centering
\footnotesize
\caption{Performance metrics for different ML models.}
\begin{tabular}{lp{0.3cm}ccc}
\toprule
\textbf{Model} & \textbf{Class} & \textbf{Precision} & \textbf{Recall} & \textbf{F1 Score} \\
\midrule
Random Forest & 0 & 0.78 & \textbf{0.95} & \textbf{0.85} \\
                            & 1 & \textbf{0.80} & 0.44 & 0.57 \\
\midrule
k-NN & 0 & 0.74 & 0.76 & 0.75 \\
              & 1 & 0.47 & 0.44 & 0.46 \\
\midrule
Gradient Boosting & 0 & 0.78 & 0.84 & 0.81 \\
                  & 1 & 0.60 & 0.50 & 0.55 \\
\midrule          
Feed-Forward Neural Network & 0 & \textbf{0.82} & 0.84 & 0.83 \\
                  & 1 & 0.65 & \textbf{0.61} & \textbf{0.63} \\
\bottomrule
\end{tabular}
\vspace{0.5cm} 
\label{tab:model_performance}
\end{table}

\subsection{Discussion}
A key contribution of this study is its focus on digestate practices in non-controlled, real-world settings, which have been under explored in remote sensing. The results demonstrate that Sentinel-2 imagery can detect digestate applications on agricultural parcels of various types, offering a scalable approach for monitoring agricultural practices. This result might enable landscape-level assessments that surpass traditional methods like farmer surveys.
Challenges were noted, particularly the fact that the application of EOM is not always observable. Further research is needed to refine spectral indices and address variables like crop growth stages, seasonal effects, and weather conditions.
ML models also showed promise, with Random Forest outperforming Gradient Boosting and k-NN in precision and F1 scores for EOM detection. The Neural Network achieved the highest score in recall.
In further iterations of this research, we plan to increase the size of the training dataset through the addition of more ground truth, as it will help in integrating better ML models for operational digestate application using free, open-acess Sentinel-2 imagery. 
Finally, on the soil health spectra, studies show that overall microbial diversity remained stable, but species richness increased with digestate and mixed fertilization. The introduction of digestate also led to the emergence of new microbial taxa associated with nutrient cycling and carbon storage, suggesting potential benefits for long-term soil health and productivity \cite{article_1}.

\section{Conclusion}

This research demonstrates the feasibility of using Sentinel-2 satellite imagery to monitor digestate application on various crops. The study explored the spectral properties of EOM applications, with specific focus on liquid digestate applied to partially vegetated soils. By leveraging spectral indices such as $EOMI_1$, $EOMI_2$, and $EOMI_3$, it was possible to track changes in reflectance following digestate applications, confirming the potential of remote sensing as a reliable tool for agricultural monitoring.
The study highlights the importance of considering different crop types, seasonal factors, and application methods in developing robust detection techniques. The promising performance of ML algorithms demonstrates their potential for accurate classification and monitoring of digestate applications at a larger scale. However, further refinement of spectral indices and ML models is necessary to improve the precision and scalability of these methods. 
Future work should focus on addressing the challenges posed by varying crop growth stages and the drying process of liquid EOMs. Additionally, the integration of other remote sensing platforms and the development of advanced algorithms could enhance the accuracy and applicability of these methods, potentially leading to the widespread use of remote sensing for monitoring agricultural practices like digestate on a regional or national scale.

\section{Acknowledgements}
This work was supported by the project "Climaca" (ID: 16196) which is carried out within the framework of the National Recovery and Resilience Plan Greece 2.0, funded by the European Union – NextGenerationEU. The authors would also like to acknowledge the Greek Payment Authority "OPEKEPE" for providing the LPIS geometries and crop type data, as well as "EPILEKTOS S.A." for their excellent collaboration and the provision of digestate application annotations.

\small
\bibliographystyle{IEEEtranN}
\bibliography{references}

\end{document}